\documentclass[lettersize,journal]{IEEEtran}
\usepackage{amsmath,amsfonts}
\usepackage{algorithmic}
\usepackage{algorithm}
\usepackage{array}
\usepackage[caption=false,font=footnotesize,labelfont=rm,textfont=rm]{subfig}
\usepackage{textcomp}
\usepackage{stfloats}
\usepackage{url}
\usepackage{verbatim}
\usepackage{graphicx}
\usepackage{cite}
\usepackage[justification=centering]{caption}
\usepackage{amsmath}
\usepackage{tabularx} 
\usepackage{makecell} 
\usepackage{xcolor}
\usepackage{multirow}

\begin{document}

\title{Physics Informed Digital Twin for RIS-assisted Wireless Communication System}

\author{Qiaojian Han,~\IEEEmembership{Student Member,~IEEE,} Haixia Zhang,~\IEEEmembership{Senior Member,~IEEE,} Yueheng Li,~\IEEEmembership{Member,~IEEE,} Dongfeng Yuan,~\IEEEmembership{Senior Member,~IEEE,} and Xiaoyan Ma~\IEEEmembership{Member,~IEEE}
\thanks{}
\thanks{Qiaojian Han is with the School of Control Science and Engineering, Shandong University, Jinan, China.

Qiaojian Han, Haixia Zhang, Yueheng Li and Dongfeng Yuan (e-mail: qjhan@mail.sdu.edu.cn; haixia.zhang@sdu.edu.cn; yueheng.li@sdu.edu.cn; dfyuan@sdu.edu.cn) are all with Shandong Key Laboratory of Intelligent Communication and Sensing-Computing Integration, Jinan 250100, China. 

Haixia Zhang and Yueheng Li are also with the Institute of Intelligent Communication Technology, Shandong University, Jinan, China.

Dongfeng Yuan is also with the School of Qilu Transportation, Shandong University, Jinan, China.

Xiaoyan Ma (e-mail: ma946@purdue.edu) is with the Purdue University,  West Lafayette, US; (The corresponding author is Haixia Zhang, e-mail: haixia.zhang@sdu.edu.cn).}}

\markboth{IEEE Wireless Communication}%
{Shell \MakeLowercase{\textit{et al.}}: A Sample Article Using IEEEtran.cls for IEEE Journals}


\maketitle

\begin{abstract}
Reconfigurable Intelligent Surface (RIS) is widely recognized as one of the key technologies for 6G due to its ability to enhance communication signal coverage and quality. To fully explore the potential of RIS, its element phase shift must be optimized. Traditional schemes dedicated for parameter design of RIS such as finite element methods (FEM) and ray-tracing, are not suitable for handling dynamic scenarios due to their high computational complexity. Therefore, digital twin is adopted, but is still far from real-time. To address this, we introduce and combine Physics Informed Neural Networks (PINN) with digital twin to build digital replicas of RIS in virtual space for dynamic channel conditions. The proposed physics informed digital twin architecture  integrates physical and data information, consisting of sensing, modeling, real-time interaction, prediction, and  phase shift optimization. In this work, the details on how to integrate PINN with digital twin to model RIS-assisted wireless communications are described, and a use case performance is analyzed. In addition, possible research directions and challenges are discussed. Simulation results show that the proposed scheme greatly reduces the prediction time compared to FEM and achieves better accuracy.
\end{abstract}


\section{Introduction}
\IEEEPARstart{W}{ith} the arrival of the big data era, a large amount of wireless communication applications requiring high reliability and high throughput have been emerging. To cater to such a trend, the future wireless communication systems expand their communication frequency to even higher frequency bands region, making wireless signals susceptible to link blocking, resulting in a degraded quality of communication and a reduced coverage. To overcome these problems, the programmable surface structure Reconfigurable Intelligent Surface (RIS) \cite{ref1,ref2} has been proposed, which can adjust the reflection of electromagnetic waves by changing the electromagnetic characteristics of the surface. Therefore, placing RIS between the transmitter and receiver can help reconstruct the physical transmission channels, making it a promising technology that can enhance the signal quality at the receiver at very low cost. 

To fully explore the potential of  RIS, its element phase shift should be carefully designed. There are schemes focusing on this topic, such as finite element methods (FEM) and ray-tracing \cite{ref3}. These schemes are typically based on theoretical models with many strong assumptions, resulting in a performance gap between theoretical derivations and real-world outcomes.  In actual communication scenarios, RIS phase shift parameters must be dynamically adjusted in real-time to accommodate varying environmental conditions. Unfortunately, the high computational complexity of those traditional  methods makes them incapable of meeting the real-time requirement. How to accurately build replicas of digital RIS system while enabling efficient real-time element phase shift optimization is still an open problem.

Digital twin \cite{ref4,ref5} is  adopted and expected to create high-quality replicas for precise virtual simulations of RIS-assisted communication systems while enabling real-time analysis and providing timely feedback to mirror real-world conditions. However, current research on digital twin \cite{ref6} is still in its early stages and faces difficulties in achieving precise modeling and real-time analysis, limiting the full realization of its potential. Therefore, a method requiring not so much data and fast computing speed is needed. Investigations show that Physics Informed Neural Network (PINN) \cite{ref7}  could be a nice candidate, since they are neural networks used to train and solve partial differential equations (PDE) by leveraging known physical laws and limited sample data. Compared to finite element methods (FEM) that rely solely on physical laws, PINN offers faster solution speeds. Additionally, by incorporating data, PINN extracts data information during training to ensure accuracy. Considering these features, PINN is highly suitable for performing real-time analysis of digital twin models for RIS in complex scenarios where data is limited and must be processed quickly. However, integrating PINN into the digital twin modeling process to assist in electromagnetic field prediction and achieve precise modeling and real-time analysis remains an area for further exploration.

\begin{figure*}[!t]
\centering
\includegraphics[width=1\linewidth]{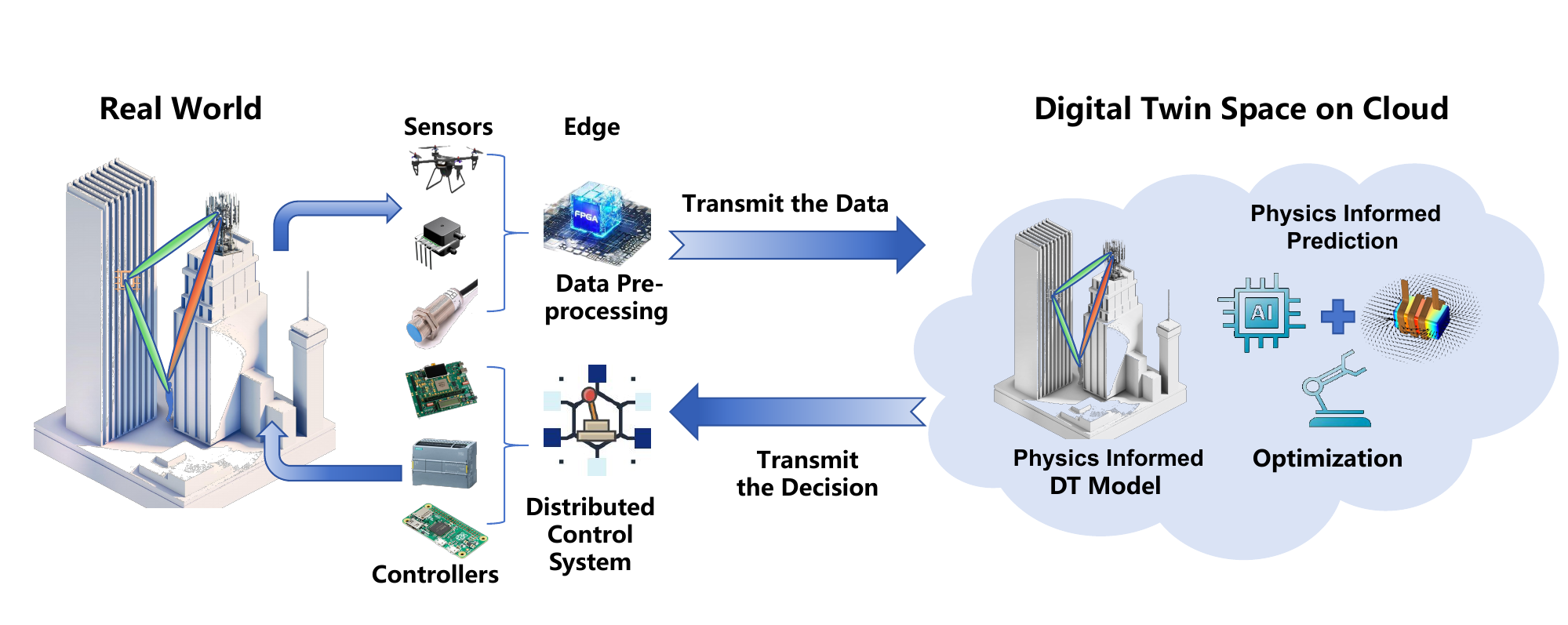}
\caption{The architecture for a physics informed digital twin for RIS-assisted wireless communications.}
\label{Fig1}
\end{figure*}

Motivated by the aforementioned features, this paper proposes a novel physics informed digital twin architecture that combines PINN with digital twin to support the implementation of RIS in wireless communication systems. This architecture includes data sensing and scene modeling, providing boundary conditions and data for PINN. As a modeling method within this architecture, PINN integrates both physical and data information to capture complex real-world variables and support pre-trained models for rapid solution generation. After solving the electromagnetic field distribution through the PINN model, the optimization component within the digital twin architecture computes the RIS phase shift results, enabling real-time feedback to the real world. Simulation results demonstrate that the proposed architecture can construct electromagnetic effect models for RIS more accurately and quickly compared to traditional methods. The model assists RIS in tracking users to achieve optimal communication performance. The paper also explores potential research directions and addresses key challenges in detail.

\section{Architecture of Physics Informed \\ Digital Twin for RIS-assisted Wireless Communication System}
This section introduces the architecture of a physics informed digital twin as  shown in Fig. \ref{Fig1} Both physical and data information are embedded in the architecture for better simulations of RIS-assisted wireless communications. The functions of each part are described, and the methods used to achieve these functions are explained in each subsection.

In a RIS-assisted communication system, data is first sensed and collected by various sensors in the real world. After preprocessing at the edge, the data is input into a digital twin space in the cloud. This digital twin includes a physics informed digital twin model, a physics informed prediction unit, and an element phase shift optimization unit. The optimized parameters are then output to a distributed control system (DCS), where the controller instructs the parameter settings in the real system.

\begin{figure*}[!t]
\centering
\includegraphics[width=0.83\linewidth]{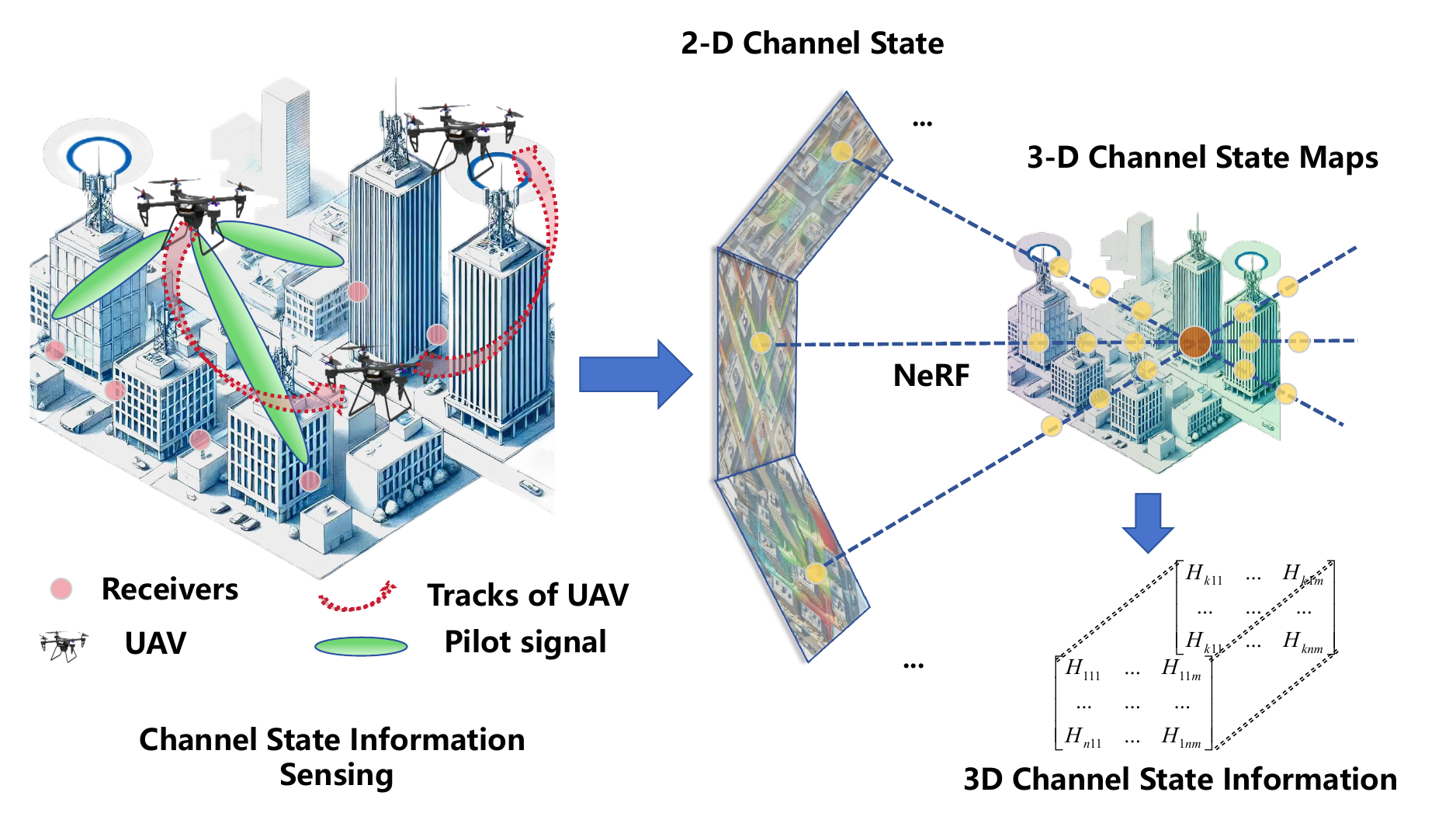}
\caption{The process from information acquisition to the 3D channel state map based on the UAV and NeRF.}
\label{Fig2}
\end{figure*}

Obviously, accurate sensing and collection of data is necessary for constructing the digital twin-driven RIS-assisted wireless communication system. Therefore, the first  step is  to sense communication scenarios, user data, and RIS characteristic data. After gathering enough data, the next step involves constructing a model to accurately replicate the real world. In digital twin, the most fundamental model is a visual mapping of the real world, which is a geometric model using 3D reconstruction. Building on the geometric model, it is necessary to understand the physical mechanisms of the existing models, especially communication-related characteristics like channel state information, the reflective or transmissive properties of objects, and the anomalous reflection characteristics of RIS. This ensures that the constructed model closely aligns with real-world conditions. The construction of these physical models requires not only the integration of physical relationships and geometric boundary conditions but also a deep fusion of real data with physical principles.

Real-time interaction between the virtual and the real world is done via collaborative communication between cloud and edge. The virtual space is placed on the cloud, and data is preprocessed at the edge. The twin in the cloud includes both the constructed models and prediction models with optimization algorithms. This facilitates intelligent decision-making. The collected data is input into the twin in real time. The resulting strategies generated by the twin are fed to the DCS, which drives the controllers to adjust the real world's RIS parameters immediately.

\subsection{Sensing of Channel and User Parameters}
In a RIS-enabled communication scenario, a base station equipped with multiple antennas communicates through an RIS composed of multiple elements. The physical channel between the base station and the user can be divided into three parts: the channel from the base station to the RIS, the reflection of the RIS, and the channel from the RIS to the user. The parameters of the base station are known, and the parameters of the RIS are adjustable. Therefore, the required information to be sensed includes the two channel segments and the user's position. Since the obstacles and other factors that significantly affect the channel are relatively fixed, the data to be collected can be categorized into persistent data and time-varying data. In details, coarse-grained channel state sensing and the geometric information for constructing the 3D model belong to persistent data, while user position and real-time channel state sensing belong to time-varying data.

Channel state information is collected by pilot-based sensing. Unmanned Aerial Vehicles (UAVs) are employed as pilot signal transmitters. A large number of receivers are uniformly deployed on the ground under the communication coverage. Upon the movement of UAVs, multiple channel state maps from different perspectives are scanned. This process produces images similar to photographs of channel state and is used to reconstruct 3D channel maps containing channel state information between reference points. Similarly, multi-angle photography is performed by UAVs to obtain data required for constructing 3D geometric models. The data encompasses not only 3D geometric information but also the physical characteristics of electromagnetic waves, thereby enhancing modeling accuracy.

For time-varying data, sensing user positions is a crucial aspect, since the real-time movement of users substantially affects the channel. As an efficient solution, RIS-related signal processing is utilized for UE position sensing, without the necessity of additional UAV efforts. By performing statistical RIS and base station beamforming over a given 3D space, the local optimum of the channel is determined, indicating the position of the receiving users. Through the arrival and departure angles at the base station and RIS, these three parts of the channel can be represented, thereby indicating the signal-to-noise ratio (SNR) \cite{ref8}. By searching different positions to find the maximum SNR, the user's location can be sensed. During the process of sensing the user's location, if the search process is too slow, it becomes difficult to ensure the real-time accuracy of user positioning. Therefore, the SNR can be established as a reward value, and this search process can be optimized through reinforcement learning \cite{ref9}. Once a user is successfully identified, their movements can be tracked to conserve sensing resources. Using the above signals, real-time channel state information is also sensed. This serves as a refinement to the previously sensed channel state.

\begin{figure*}[!t]
\centering
\includegraphics[width=0.75\linewidth]{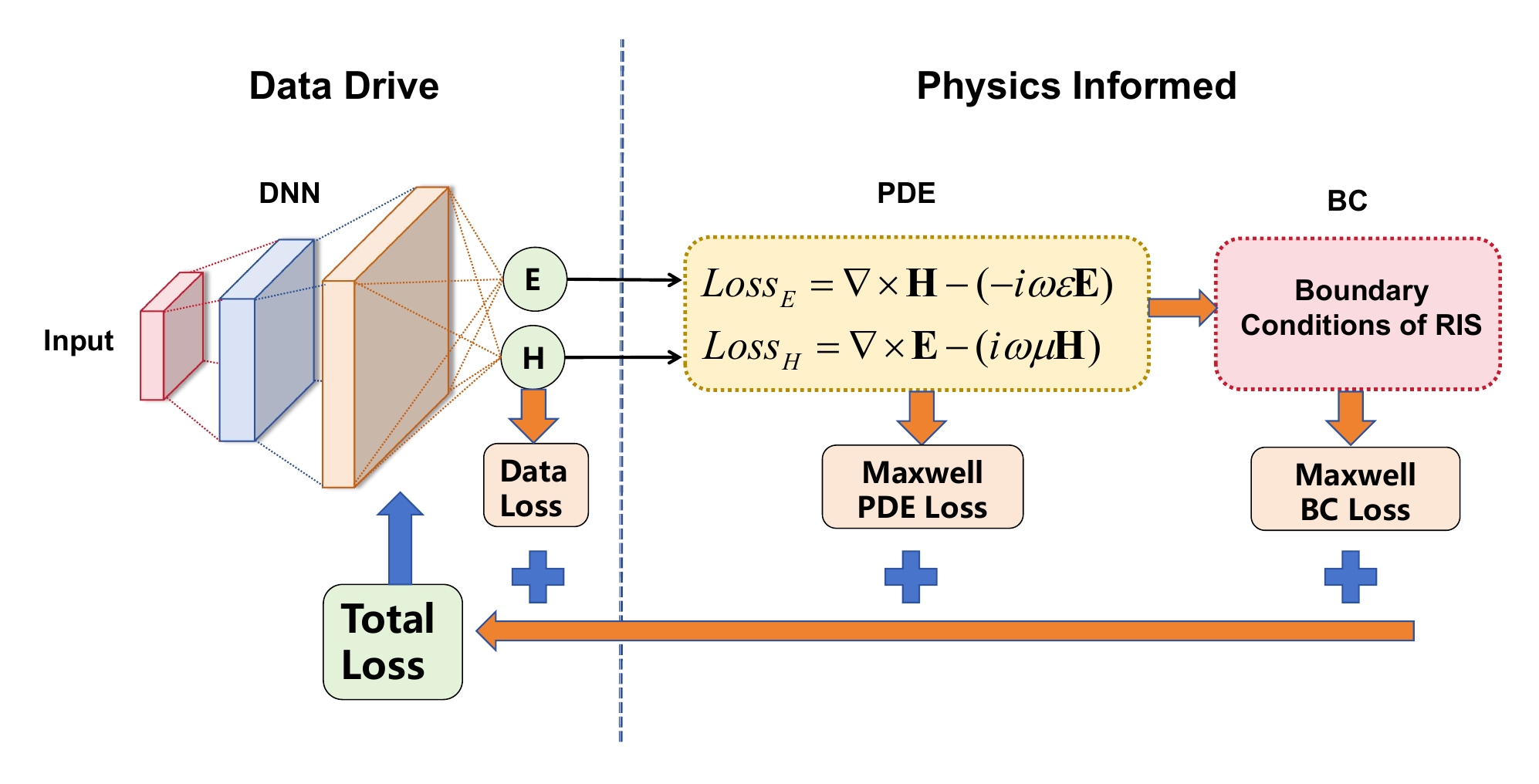}
\caption{The PINN framework for solving RIS electromagnetic wave reflection properties.}
\label{Fig3}
\end{figure*}
\subsection{The Modeling of Digital Twin System}
Once sufficient data is obtained, it is important to integrate and fuse this data to create a virtual replica of the real space. The digital twin model should not only accurately reflect the visual features of the scene but also include its physical characteristics. Previous stages of digital twin primarily used traditional Computer Aided Design (CAD) and Computer Aided Engineering (CAE) tools for modeling. Although these approaches allow for precise modeling, they are time-consuming and labor-intensive, making it challenging to apply them to the geometric modeling of large-scale communication scenarios.  Therefore, adopting AI technologies such as Neural Radiance Fields (NeRF)\cite{ref10} or 3D Gaussian splatting\cite{ref11} represents a new trend for geometric modeling. These technologies enable the conversion of two-dimensional data captured through multi-perspective UAV oblique photography into 3D information.

During the modeling process, due to limited computing resources, important aspects that affect system performance should be prioritized for detailed modeling. Other parts can be modeled with basic geometry. In this communication scenario, the channel state and the RIS require precise modeling. The following sections will provide detailed descriptions of these two components.

Although NeRF and 3D Gaussian splatting have faster computation speed and more accurate real-time prediction results, making RIS configuration more precise and timely. While traditional methods are limited to geometric model construction, the way they embed physical information offers important insights for more accurate modeling. NeRF and 3D Gaussian splatting work by analyzing multiple two-dimensional pictures of an object from different angles, helping to understand the shape and appearance of the object. Then, they build a 3D representation by learning how light interacts with the surfaces in the images.

When reconstructing channel information, the light information in NeRF can be replaced with the physical quantity of channel state. Similarly, the two-dimensional channel state maps from different perspectives collected during data acquisition can be converted into 3D scene channel distributions using the concept of NeRF. This is achieved by replacing the light radiance information in NeRF with channel state information. In this way, a 3D channel map can be modeled. This process is shown in Fig. \ref{Fig2}. The constructed 3D channel map can quickly provide prior knowledge of channels for the actual communication system, ensuring a real-time feedback of the digital twin system.

After the signal transmission path is determined, the channel characteristics can be obtained based on the channel map. Then,  the compressive sensing method is applied for channel estimation based on the pilot signals. In this method, the sparsity characteristic of signals in the angular domain \cite{ref12} is utilized. On the basis of the constructed channel model, the real-time channel can be fine-tuned.

Next, we discuss the modeling of the RIS. The reason for using RIS in communication scenarios lies in the capability of this metamaterial to instantly change the phase of electromagnetic waves upon reflection at specific frequencies, meeting the required direction of departure (DoD) and direction of arrival (DoA). Therefore, the reflective properties of RIS, also known as anomalous reflection, do not follow conventional specular reflection and are more complex than those of ordinary objects.
To address this, PINN is introduced. A 3D grid region is created, where the RIS is placed. The direction and polarization of the incoming electromagnetic waves are set. In this space, the Maxwell’s equations are solved using the PINN. The detailed process is illustrated in Fig. \ref{Fig3} and described below. 

First, a limited 3D space is divided into grids, where the physical quantities of electric field strength and magnetic field strength are located within each grid cell. Real data is input into a deep neural network (DNN), and the network outputs the electric field strength and magnetic field strength in space. Since the space has been divided into grids, these outputs are in the form of 3D matrices. The residuals between these outputs and the real data form the data loss. This loss is then input into the frequency domain Maxwell’s PDE. By combining the boundary conditions from RIS, such as surface current density and abnormal reflection phase gradients, the loss from the physically constrained PDE and the boundary condition loss are calculated through weighted summation to obtain the total loss. The total loss guides the training of the network until convergence is achieved. During the training process, the electric and magnetic field residuals are calculated through forward propagation. The loss function is solved and optimized through back propagation. The program monitors the loss changes during training and applies boundary conditions to simulate reflection phenomena. Finally, the scattered electric and magnetic fields are output to describe the direction of reflection. Therefore, after the training of the PINN model is completed, its application for predictive calculations is significantly faster than FEM.

Based on the constructed models with physical mechanisms, the operation of the  RIS-assisted communication system can be simulated and rehearsed in virtual space in advance. In this virtual space, actions that are difficult to execute in the real world can be simulated. These include moving RIS, adjusting their angles, repositioning base stations, and altering signal strengths. This simulation allows us to study how such changes affect the performance of communication systems.

\subsection{Real-time interaction of Digital Twin System}
Real-time data interactions are key points where digital twin differs from traditional methods. Efficient data interaction and processing mechanisms should be developed to establish a highly integrated and real-time synchronized system.

To begin with, a three-dimensional geometric scene is established using a digital twin virtual platform. The algorithms describing the communication system are implemented in the C++. These are organized into classes and then integrated with the 3D geometric scenes. A two-directional data channel between the virtual and real environments is established using Socket or other communication protocols. To balance computation and communication resources in the digital twin system, edge devices preprocess sensor data before uploading them to the cloud-based virtual digital twin space via Socket. Similarly, the feedback and decisions are transmitted to the edge through the cloud's digital twin system, and the controllers are driven accordingly. Through this process, the digital twin system ensures real-time data acquisition and exchange with the external world, achieving system-wide real-time interaction.

\subsection{Optimization of RIS phase parameters in digital twin System}
Optimization of RIS phase parameters in the digital twin system is the main goal of building the digital twin. After solving the channel parameters in the environment using prediction methods, the optimization of RIS phase parameters can be obtained in advance in the virtual space based on the predicted channel state information. This approach accelerates the setup of RIS parameters in the actual system.

The prediction function utilizes real-time and historical information to predict and infer future scenarios. This includes predicting user locations, forecasting channel conditions, and deducing electromagnetic characteristics for optimization purposes.
A time-domain and full-scene PINN is designed to solve time-varying channel state information and other related parameters to establish a simulation framework for the communication. The real-time sensed user position information is used as the initial condition for the network. The constructed 3D channel model serves as the boundary conditions. The channel state information is set as the target variable to be solved. The trained model can predict the channel state for the coming time slot.

\renewcommand{\arraystretch}{1.4} 

\begin{table*}[htbp]
\centering
\caption{Comparisons of Computation Time, Accuracy, and Impact on Communication System for PINN and Other Methods}
\resizebox{\textwidth}{!}{ 
\begin{tabular}{l|cc|ccc|cc}
\hline
\multirow{2}{*}{Method} & \multicolumn{2}{c|}{Computation Time} & \multicolumn{3}{c|}{Computation Accuracy} & \multicolumn{2}{c}{Impact on Communication System} \\
\cline{2-8} 
 & Training Time & Prediction Time & L2 Residual & RMSE\_E & MAE\_E & Reflection Efficiency & Transmission Rate \\
\hline
FEM & - & 10986.7s & 0.000177 & 0.137 & 0.154 & 0.2816 & 4.800 bit/s \\
Physics Driven & 2076.6s & 3.84s &  0.007197 & 0.282 & 0.316 & 0.1567 & 3.294 bit/s \\
Data Driven & 1013.6s & 1.78s & - & 0.0564 & 0.106 & 0.4253 & 6.041 bit/s \\
\textbf{Physics Informed} & 2467.0s & 4.12s & - & 0.0028 & 0.0046 & 0.4639 & 6.289 bit/s\\
\hline
\end{tabular}
}
\label{TableI}
\end{table*}

Based on the prediction results, the system makes optimization decisions in advance, which are then fed back to the real world. The RIS parameter configurations are primarily optimized and adjusted. To ensure fairness, the objective of optimizing the phase shifts of RIS elements is to maximize the minimum downlink rate among  users. By optimizing the phase shift of each element on the RIS under constraints like sensing capacity, and total power, the maximum achievable rate for the user with the lowest rate is calculated \cite{ref13}. The optimized strategy is then transmitted in real time back to the controllers in the actual RIS communication system driving changes in the phase shifts of the RIS. In solving the optimization strategy, we proposed a reinforcement learning algorithm.

The optimal parameters are processed by the DCS and controllers. This prompts the RIS in the actual system to adjust its parameters accordingly, leading to an improvement in communication quality.

\begin{figure}[htbp]
\centering

\subfloat[The 3D streamlines of FEM]{
\includegraphics[scale=0.15]{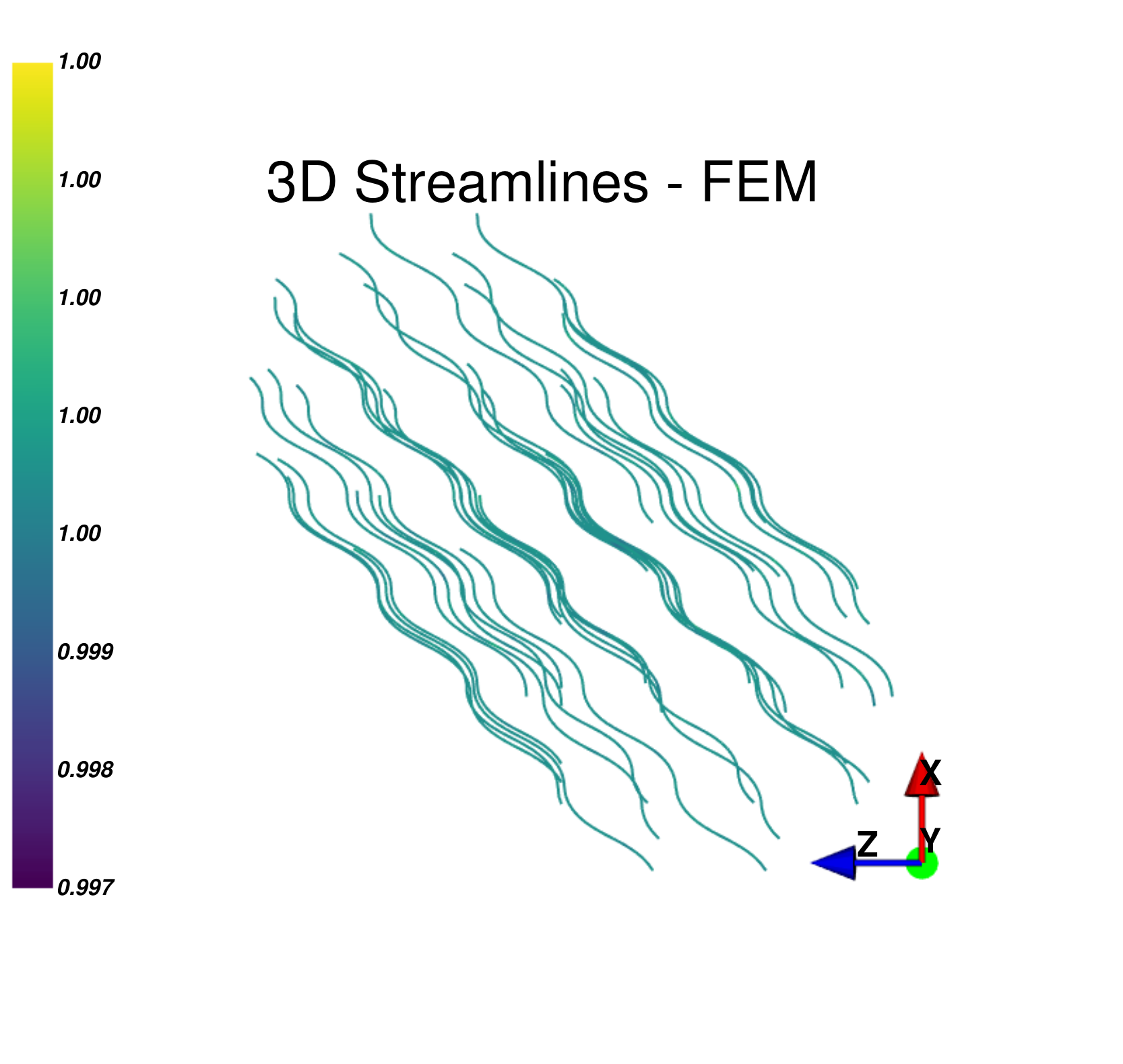}
\label{Fig4(a)}}
\subfloat[The 3D streamlines of PINN]{
\includegraphics[scale=0.15]{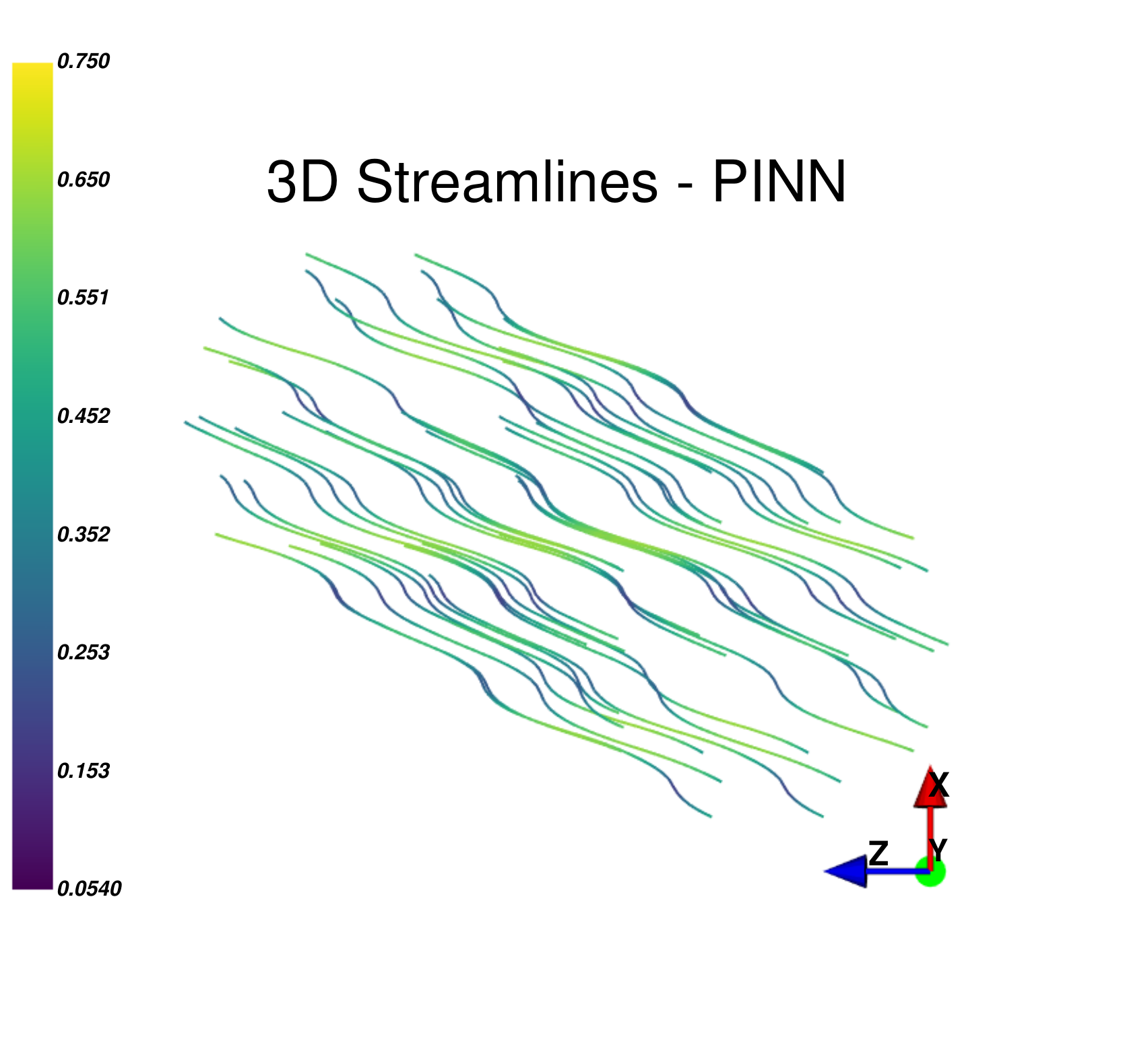}
\label{Fig4(b)}}

\caption{The result of FEM and PINN shown by the electric field 3D streamlines}
\label{Fig4}
\end{figure}

\section{Experimental Results and Analysis}
We first construct a preliminary RIS model including RIS itself and its characteristics of wave reflection driven by PINN, by integrating physical features and data. This model is used as a case study to demonstrate the role of PINN in physics informed digital twin. We show the anomalous reflection of the RIS under vertical incidence in Fig. \ref{Fig4}\subref{Fig4(a)} and \ref{Fig4}\subref{Fig4(b)}, respectively. The results are indicated by the electric field 3D Streamlines, which represents the direction of electric field.

To highlight the benefits of the proposed physics informed approach, FEM and PINN are compared in three aspects as shown in Table \ref{TableI}. Both PINN and FEM are configured with identical simulation settings on a computational domain consisting of 150×150×150 units. Both the FEM solving iterations and PINN training epochs are set to 500, and all computations are conducted on an NVIDIA Tesla V100 GPU. Firstly, as shown in Table \ref{TableI} although PINN requires additional training time, but its prediction time is much shorter than FEM, which makes it suitable to real-time applications. In terms of accuracy, the L2 residual indicates that FEM conforms better to the underlying physics under ideal conditions. However, PINN achieves lower RMSE and MAE, leading to more precise electric field predictions in real-world scenarios, and thus more accurate predicted channel state information. Finally, Table \ref{TableI} shows that the communication performance of these schemes, the PINN-based approach attains higher reflection efficiency and transmission rates compared to FEM, highlighting its advantage in optimizing reflection control for wireless systems.

\begin{figure*}[htbp]
\centering
\subfloat[An overhead view of the system]{
\includegraphics[scale=0.28]{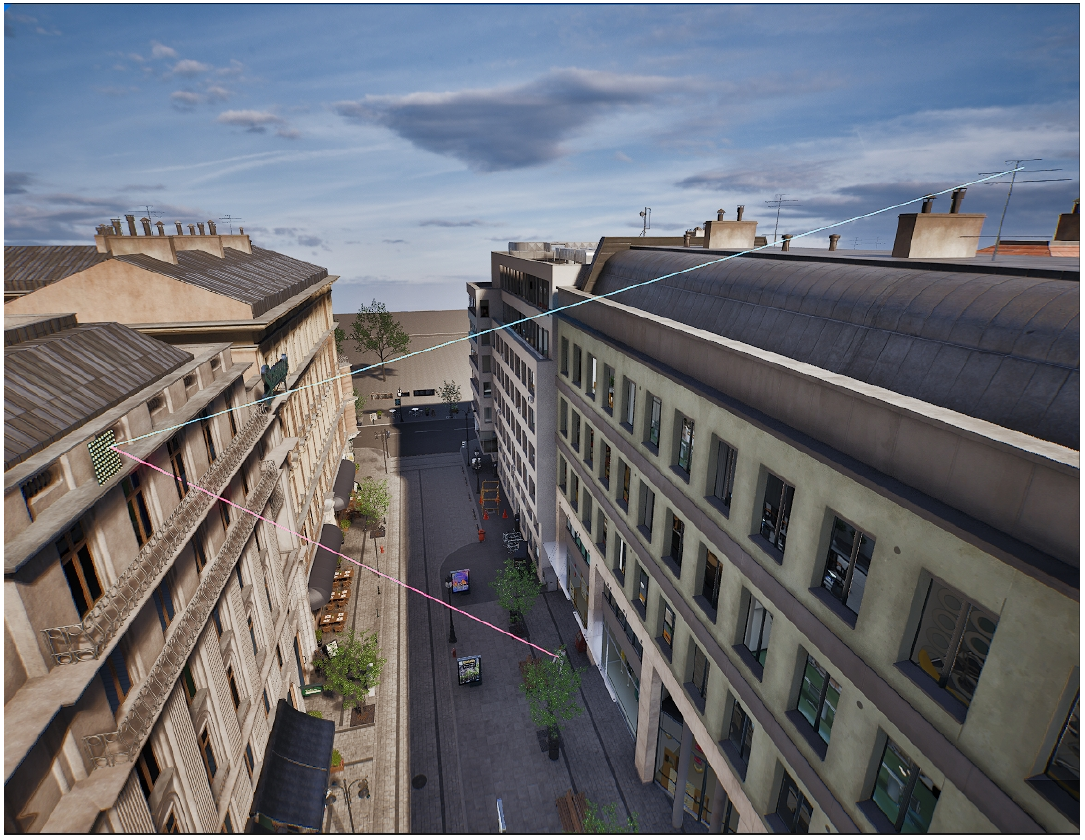}
\label{Fig5(a)}}
\subfloat[The RIS in the digital twin system]{
\includegraphics[scale=0.23]{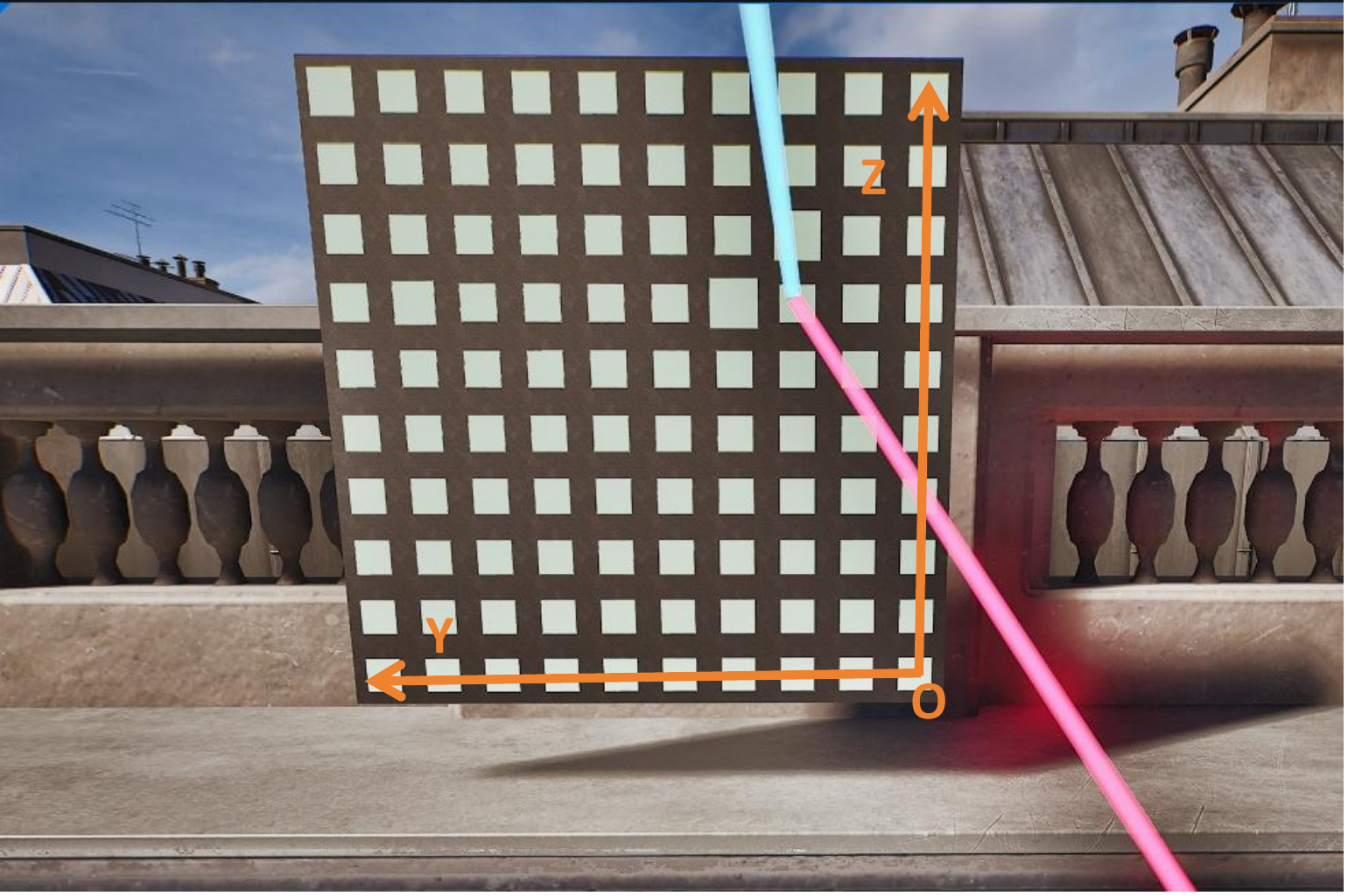}
\label{Fig5(b)}}
\subfloat[The user in the digital twin system]{
\includegraphics[scale=0.28]{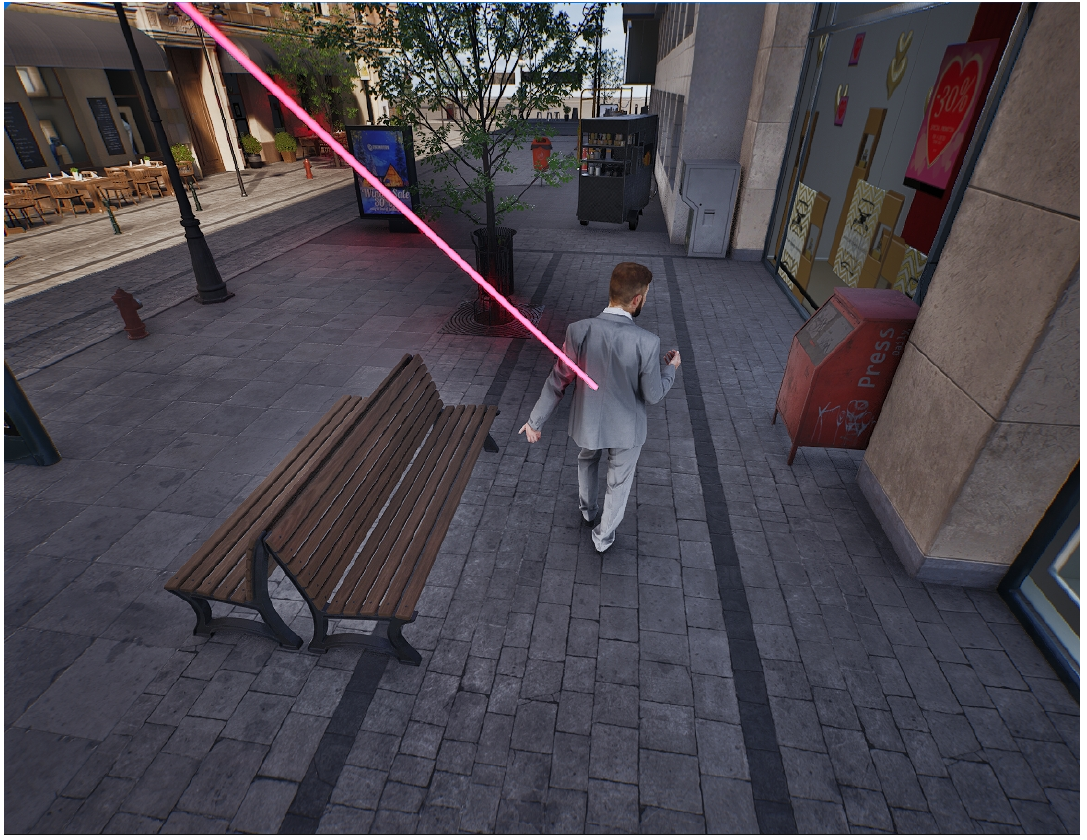}
\label{Fig5(c)}}
\qquad
\subfloat[The anomalous reflection angles and their corresponding phase differences at each time.]{
\includegraphics[scale=0.4]{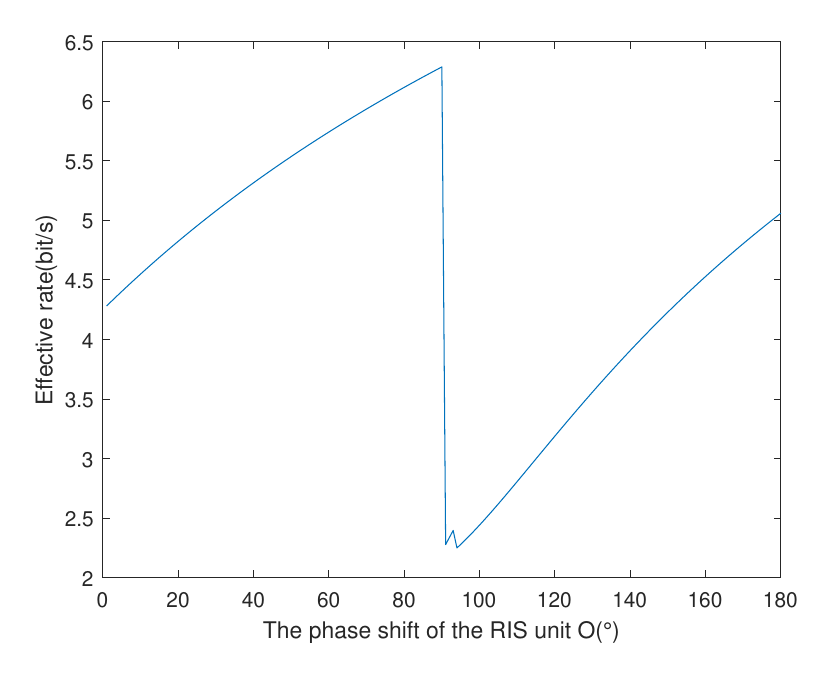}
\label{Fig5(d)}}
\subfloat[Transmission rate of user as a function of the phase shift of RIS unit O.]{
\includegraphics[scale=0.4]{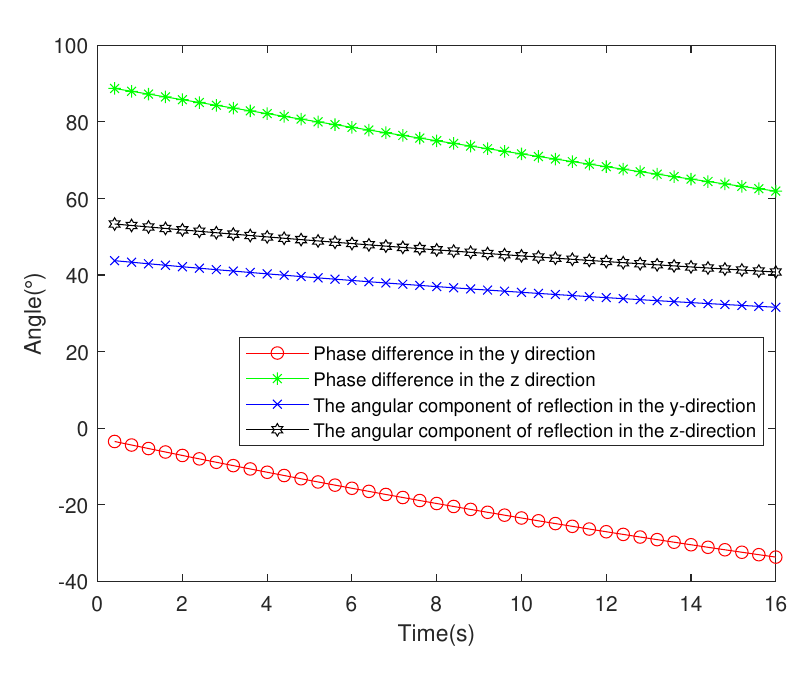}
\label{Fig5(e)}}

\caption{The constructed digital twin system for RIS-assisted wireless communication systems}
\label{Fig5}
\end{figure*}

We further construct a digital twin system for RIS-assisted wireless communication system to obtain the parameters of RIS units and the corresponding system performance in terms of the transmission rate as users move. The setup utilizes Archmodels for UE vol. 8 geometric model data, and for some data not actually collected, such as channel state information, a used Rician fading channel model is adopted considering that the RIS is deployed on the wall of tall buildings. During experiments, the electromagnetic wave width of the communication signal is configured, using RIS units for reflection. An overhead view of the system is shown in Fig. \ref{Fig5}\subref{Fig5(a)}. Fig. \ref{Fig5}\subref{Fig5(b)} shows the RIS in the virtual system, and Fig.  \ref{Fig5}\subref{Fig5(c)} displays an image at the user side. Digital twin visualizations from three views show the communication link from the base station to the RIS and then to the user. The methods described in Section II are all embedded in the system. Fig.  \ref{Fig5}\subref{Fig5(d)} shows how the phase difference of RIS units varies to keep the main lobe reflection aligned with the moving user at a constant speed. We define that the main lobe reflection is aligned with the location of user if the received signal power at the user side reaches its maximum. Additionally, the phase difference is defined as the difference in phase shifts between neighboring RIS units along the positive y and z directions in Fig.  \ref{Fig5}\subref{Fig5(b)}. This result confirms that phase difference affects the angular component of reflection and displays the optimal phase difference parameters of the RIS at each moment. In reality, the transmission rate is a function of the phase shift matrix consisting of all RIS units. The phase difference in  \ref{Fig5}\subref{Fig5(d)} and the phase shift in  \ref{Fig5}\subref{Fig5(e)} jointly determine the phase shift matrix of all RIS units, thereby determining the transmission rate. This also confirms that the optimal phase parameter for all RIS units can then be calculated based on the optimal phase differences and optimal phase of Fig.  \ref{Fig5}\subref{Fig5(e)}. By determining the optimal parameters for the signal to follow the user in real-time, these parameters are fed back to the real world, enabling the digital twin framework to guide real-world parameter configurations.

These experiments validate the feasibility and effectiveness of integrating PINN into the digital twin system for RIS-assisted wireless communications. By accurately modeling the physical behaviors and dynamically optimizing the system parameters, the physics informed digital twin provides a powerful tool for improving communication performance in complex and dynamic environments.

\section{Possible Research Directions and Challenges}
The development of the physics informed digital twin for RIS-assisted wireless communication systems presents numerous challenges and opens up various research directions. This section discusses key challenges and potential solutions, aligning with cutting-edge technologies to advance the field.

\subsection{Integrated Communication-Sensing-Computing and Digital Twin}
The implementation and operation of digital twin systems require substantial computing, communication, and sensing resources, which in resource-constrained applications can impact existing communication systems and possibly reduce the overall system efficiency. To address this challenge, a novel technology known as integrated communication-sensing-computing is required\cite{ref14}. This technology focuses not only on the effective allocation of resources but also on achieving dynamic optimization and efficient configuration of resources through the deep integration of communication, sensing, and computing functions. Effective allocation of resources between the existing communication system and the digital twin system ensures the normal operation of the former and the quality of the latter. Specifically, research can be conducted from the following aspects:
\begin{itemize}
    \item \textit{Task prioritization}: Task prioritization ensures that critical tasks have priority to access resources, maintaining overall system stability and efficiency. Specifically, tasks can be categorized into high-priority and low-priority groups according to their characteristics. For example, phase adjustment and real-time electromagnetic field computation are considered of high-priority tasks, while model training and updating are classified to be of low-priority. By assigning priority weights and employing a priority scheduling algorithm to determine the execution location and queue position of each task, the system ensures that overall Quality of Service (QoS) requirements are met.
    \item \textit{Network slicing}:  Digital twin, sensing, and communication tasks all consume resources, resulting in resource competition within the system. This competition may affect the performance of high-priority tasks. Network slicing can be used to allocate dedicated resources for different tasks or services, ensuring their QoS requirements. Slices should be defined based on task requirements, categorizing tasks into real-time and low-priority slices. When the number of real-time tasks increases, the bandwidth allocated to low-priority slices can be temporarily reduced to prioritize real-time QoS. Machine learning algorithms can be applied to predict future task demands, allowing resources to be allocated in advance.
    \item  \textit{Intelligent resource management}: Integrating artificial intelligence technology into resource management can predict system demands and automatically adjust resource configurations to adapt to real-time changes in the system. Reinforcement learning and other AI-based algorithms can be applied to jointly design multi-objective optimization strategies. These models can balance communication, sensing, and computing resources to achieve long-term best QoS.
\end{itemize}

\subsection{Real-time and Parallel Computing of Digital Twin}
The biggest challenge in synchronizing virtual and real worlds in a digital twin system is real-time computing. The accuracy of the virtual space relies on many artificial intelligence algorithms for calculations. Achieving real-time and parallel computing while ensuring calculation accuracy is a major challenge for future research.

\begin{itemize}
    \item  \textit{Boundary conditions for Physics Informed Digital Twin}: 
In some scenarios, boundary conditions may change over time. How to effectively capturing and updating these dynamic boundary conditions remains an open problem. For real datasets, boundary conditions are often affected by noise. How to address the uncertainty caused by noise to improve the robustness of physics informed digital twin is another issue to be explored.

    \item \textit{The parallel computation of Physics Informed Digital Twin}: 
Current PINN models still struggle to process temporal information effectively. Although they are much faster than FEM, their computational capacity is still not efficient enough for complex digital twin scenarios. Additionally, improving and coordinating the computing ability of various AI models involved, such as PINN and reinforcement learning, are necessary. It could be benificial if we can create a unified model or generative AI \cite{ref15} applicable to multiple scenarios for digital twin applications in communications.
    \item  
 \textit{The coupling of macro and micro phenomena and multiple physical fields}: 
The differences between macro and micro phenomena complicate the construction and solution of physical models. Different scales of physical processes require distinct descriptive methods. How to effectively combine these methods is crucial. The coupling of multiple physical fields requires the simultaneous consideration of multiple equations, which increases the computational complexity. Interactions between different physical fields may involve complex nonlinear effects, making the models difficult to solve.
\end{itemize}

\section{Conclusions}
This paper proposes an architecture using a physics informed digital twin for RIS-assisted wireless communication systems. The architecture is explained from several aspects, including sensing, modeling, real-time interaction, prediction, and element parameters optimization. Experimental results show the effectiveness of integrating physics informed neural networks into the digital twin system in speeding up the phase shift optimization. Additionally, future research directions and challenges in this area are identified and discussed. It is expected that this work will provide a novel approach and framework for physics informed digital twin to facilitate the parameter design of RIS-assisted wireless communication systems. This facilitates the realization of self-optimizing, self-configuring, and self-healing RIS-assisted wireless communication systems and promotes its development in 6G communications.

\section{Acknowledgment}
This work was supported in part by the National Natural Science Foundation of China under Grant U22A2003, Grant U23A20277, Grant 62271288, the Mobility Programme of Sino-German Cooperation and Exchange NSFC under Grant M-0626, the Key R\&D Program of Shandong Province, China under Grant No.2022CXGC020107, and the First Seedling Program of Shandong University under Grant SDU-QM-Y2024009.

\section{Biographies}
Qiaojian Han [S’23] (qjhan@mail.sdu.edu.cn) is currently pursuing his Ph.D. degree at Shandong University. His research interests include reconfigurable intelligent surface, digital twin enabled wireless communication and digital twin in IIoT.

Haixia Zhang [M’08-SM’11] (haixia.zhang@sdu.edu.cn) received the B.E. degree from the Department of Communication and Information Engineering, Guilin University of Electronic Technology, China, in 2001, and the M.Eng. and Ph.D. degrees in communication and information systems from the School of Information Science and Engineering, Shandong University, China, in 2004
and 2008, respectively. From 2006 to 2008, she was an Academic Assistant at the Institute for Circuit and Signal Processing, Munich University of Technology. From 2016 to 2017, she was a Visiting Professor with the University of Florida, USA. She is currently a Distinguished Professor with Shandong University. Her current research interests include industrial Internet of Things (IIoT), resource management, mobile edge computing, and smart communication technologies. She serves as a TPC member, a session chair, an invited speaker and a keynote speaker for conferences. She is on the Editorial Boards of the IEEE Transactions on Wireless Communications, IEEE Internet of Things Journal, IEEE Wireless Communication Letters, and China Communications.

Yueheng Li [S’18-S’23] (yueheng.li@sdu.edu.cn) Yueheng Li is currently a research fellow at the Institute of Intelligent Communication Technology, Shandong University. His research interests include reconfigurable intelligent surface, 6G system design, and integrated sensing and communication.

Dongfeng Yuan [SM’01] (dfyuan@sdu.edu.cn) is a full professor with Shandong University, China. From 1993 to 1994, he was with the Electrical and Computer Department at the University of Calgary, Alberta, Canada. He was with the Department of Electrical Engineeringin the University of Erlangen, Germany, from 1998 to 1999; the Department of Electrical Engineering and Computer Science in the University of Michigan, Ann Arbor, USA, from 2001 to 2002; the Department of Electrical Engineering in Munich University of Technology, Germany, in 2005; and the Department of Electrical Engineering Heriot-Watt University, UK, in 2006. His current research interests include digital twin, IIoT and cooperative (relay) communications.

Xiaoyan Ma [M’23] (ma946@purdue.edu) received the B.S. and Ph.D. degrees from the School of Information Science and Engineering, Shandong University, Jinan, China, in 2017 and 2023, respectively. From August 2023 to December 2024, she worked as a Research Fellow with the School of Electrical and Electronic Engineering, Nanyang Technological University, Singapore. From Janauary 2025, she started to purshing her second PhD degree in Purdue University, West Lafayette, US. Her research interests include multiple-in–multiple-out technology, reconfigurable intelligent surface, orbital angular momentum waveform design, and federated learning.

\vfill

\end{document}